\documentclass[10pt,twocolumn,letterpaper]{article}

\usepackage[pagenumbers]{cvpr}   %

\usepackage{array}
\usepackage{float}
\usepackage{algorithm}
\usepackage{algpseudocode}
\usepackage{multirow}
\usepackage{listings}
\usepackage{url}

\newif\ifpreprint
\newcommand{\otherref}[2]{\ifpreprint\ref{#1}\else #2\fi}
\newcommand{\ofsupp}{\ifpreprint\else\ of the supplementary\fi}

\newcommand{\examplecaption}{A \dataset{} question, shortened to four hops. Appendix~\otherref{app:corpus}{F}\ofsupp{} gives the full question.}

\providecommand{\nolinenumbers}{}
\providecommand{\linenumbers}{}

\newcommand{\dataset}{Video-HopChain}
\newcommand{\method}{Confidence-Gated Exploration}
\newcommand{\methodshort}{CGE}
\newcommand{\model}{Video-HopChain-8B}
\usepackage{colortbl}
\definecolor{ourrow}{HTML}{DCE9F8}
\newcommand{\ourrow}{\rowcolor{ourrow}}
\definecolor{hopA}{HTML}{3F74AB}
\definecolor{hopB}{HTML}{7B62A8}
\definecolor{hopC}{HTML}{2F8F86}
\definecolor{hopD}{HTML}{B8693F}

\newcommand{\E}{\mathbb{E}}
\newcommand{\pstar}{p^{\star}}
\newcommand{\ystar}{y^{\star}}

\newcommand{\clip}{\operatorname{clip}}

\usepackage{microtype}
\preprinttrue

\definecolor{cvprblue}{rgb}{0.21,0.49,0.74}
\usepackage[pagebackref,breaklinks,colorlinks,allcolors=cvprblue]{hyperref}
\hypersetup{pdftitle={Video-HopChain: Multi-Hop Questions and Confidence-Gated Exploration for Video Reasoning Models}}

\def\confName{CVPR}
\def\confYear{2027}

\newif\ifwithsupp
\withsupptrue

\newcommand{\pageurl}{https://ngquangtrung57.github.io/video-hopchain-page/}
\newcommand{\dataurl}{https://huggingface.co/datasets/ngqtrung/Video-HopChain}
\newcommand{\modelurl}{https://huggingface.co/ngqtrung/Video-HopChain-8B}
\newcommand{\codeurl}{https://github.com/ngquangtrung57/video-hopchain}
\newcommand{\baseurl}{https://huggingface.co/ngqtrung/Video-HopChain-8B-Standard-RL}
\newcommand{\collectionurl}{https://huggingface.co/collections/ngqtrung/video-hopchain}
\newcommand{\releaseurl}[1]{{\Urlmuskip=0mu\relax\url{#1}}}

\newcommand{\aff}[1]{\textsuperscript{#1}}
\usepackage{marvosym}
\newcommand{\corr}{\textsuperscript{\,\Letter}}

\title{\dataset{}: Multi-Hop Questions and Confidence-Gated Exploration for Video Reasoning Models}

\author{%
Nguyen Quang Trung\aff{1}\quad
Yuhao Dong\aff{1}\quad
Shuo Sun\aff{2}\quad
Shuai Liu\aff{1}\quad
Shulin Tian\aff{1}\\
Kim-Hui Yap\aff{3}\quad
Ziwei Liu\aff{1}\corr\\[3pt]
{\normalsize\aff{1}S-Lab, Nanyang Technological University (NTU)\qquad
\aff{2}Johns Hopkins University\qquad
\aff{3}NTU}\\[1pt]
{\small\corr\,Corresponding author}\\[4pt]
{\small\href{\pageurl}{Project page}\enspace$\cdot$\enspace
\href{\codeurl}{Code}\enspace$\cdot$\enspace
\href{\collectionurl}{Dataset and models}}%
}

\begin{document}
\twocolumn[{%
\renewcommand\twocolumn[1][]{#1}%
\maketitle
\vspace{-14pt}
\begin{center}
\includegraphics[width=0.95\linewidth]{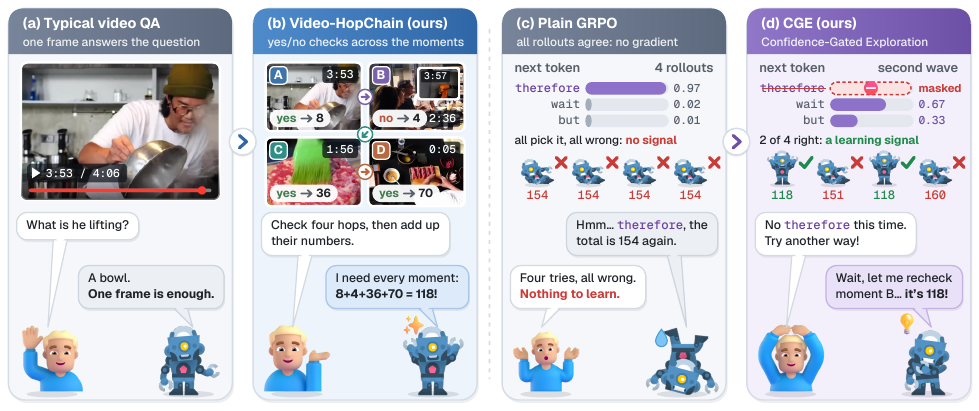}
\captionof{figure}{\dataset{} and \methodshort{} at a glance. (a, b) A typical video question needs one frame, while a \dataset{} question chains yes/no checks across the moments of a video and sums the numbers they select. (c, d) When every rollout of a group fails in the same way, the group gives no gradient. In the second wave, \methodshort{} masks the token that the policy is sure of, so the rollouts take new paths and the rewards differ.}
\label{fig:teaser}
\end{center}
\vspace{4pt}
}]

\begin{abstract}
Answering a question about a video often requires evidence from several moments, and long chain-of-thought reasoning over such evidence exposes perception and reasoning errors that compound across steps. However, most video data used for reinforcement learning with verifiable rewards (RLVR) rarely demands a chain of visual evidence, so these errors stay unexposed. HopChain addressed this problem on still images by synthesizing multi-hop questions, and we extend it to video with \dataset{}, a dataset of 22,550 multi-hop questions over 13,378 videos, together with a held-out benchmark of 1,000 questions. Each question chains three to six hops, and each hop checks one piece of visual evidence in the video. In many questions, the answer to an earlier hop decides which moment a later hop inspects, so the model must answer the hops in order. The outcome of each hop selects an integer, and the final answer is the sum of these integers, a single number that an exact match can verify. We train Qwen3-VL-8B with GRPO first on standard video data and then on \dataset{}, and this second stage raises the mean over eight video understanding and reasoning benchmarks from 55.4 to 57.9 and improves every one of them. Harder data, however, also exposes a known limitation of GRPO: a group whose rollouts are all correct or all incorrect has zero reward variance and therefore gives no gradient. To recover these zero-variance groups at no extra rollout cost, we introduce \method{} (\methodshort{}). \methodshort{} samples the first 4 of 8 rollouts as usual, and if these 4 are all correct or all incorrect, it samples the last 4 with the policy's most confident token masked inside the reasoning span. \methodshort{} raises the mean further to 59.3. \ifpreprint We release\else We will release\fi{} our model, \model{}, together with the dataset, the data generation code, and the training code.
\label{sec:abstract:end}

\end{abstract}

\begin{figure*}[t]
\centering
\includegraphics[width=0.9\linewidth]{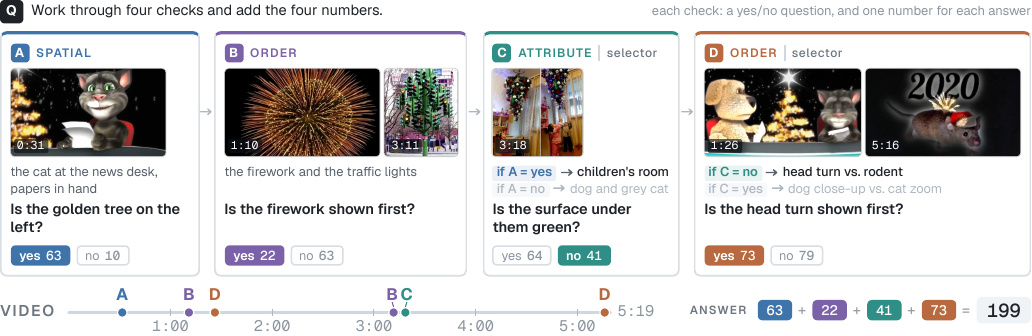}
\caption{\examplecaption}
\label{fig:example}
\end{figure*}

\section{Introduction}
\label{sec:intro}

Rule-based reinforcement learning with verifiable rewards (RLVR) has become the standard route to a reasoning model, first in mathematics and code \citep{deepseekr12025, grpo2024, yu2025dapo} and then in images \citep{openmmreasoner2025, thinklitevl2025}. Video is the natural next domain, and a growing number of systems now train video question answering models with Group Relative Policy Optimization (GRPO) against a verifiable answer \citep{videor12025, videochatr1, videorft, longvilar1, onethinker2025}. However, what such a model learns depends on the questions it is trained on. HopChain \citep{hopchain2026} made this point for still images: long chain-of-thought reasoning exposes perception, reasoning, knowledge and hallucination errors that compound across intermediate steps, yet most RLVR data asks for no reasoning chain that relies on visual evidence throughout, so these weaknesses stay largely unexposed during training. We observe the same problem in video. Much of the existing video RLVR data is already verifiable, because it is multiple choice or numeric, but the questions that would require a chain of visual evidence are often conceptual, and they are therefore hard to verify. Prior audits also report that video models often answer without looking at the video at all \citep{vidground2026, mvp2025}, and that a model can encode what it sees and still answer from its priors \citep{sensesgap2026}. We therefore design the training data so that the model must look at the video, and look at it several times, before it answers (Figure~\ref{fig:teaser}).

HopChain addresses this weakness on images with multi-hop data synthesis. It builds each question as a chain of dependent hops, in which earlier hops fix the objects and conditions that later hops need, and it ends the chain in one unambiguous number, so a wrong step changes the final answer. In addition, training on such data yields reasoning that generalizes across general understanding and reasoning benchmarks. We expect such chains to matter most in video, because the evidence a chain must revisit is spread over time rather than held within a single frame.

For this reason, we adapt this framework to video and build \dataset{}, in which every question asks about the moments of one video rather than about the regions of one image. Each question chains three to six hops, and every hop checks one piece of visual evidence at one or two moments of the video, with an answer that selects one of two integers, so the final answer is the sum of these integers and is verifiable by exact match. Almost every question includes at least one hop on the order of two moments, so a single frame does not carry the evidence that a chain needs. In addition, some hops act as selectors, where an earlier answer decides which moment a later hop examines. To generate the questions, we first caption each shot of a video with a vision-language model, and then we use a text-only language model to write the multi-hop question from those captions. The resulting dataset holds 22,550 training questions over 13,378 videos, plus 1,000 held-out videos with one question each as a benchmark.

We expect that training on this dataset improves general video understanding and reasoning, even though the dataset targets one specific reasoning type. To test this, we train Qwen3-VL-8B-Instruct with GRPO and evaluate on eight general-purpose video benchmarks. Table~\ref{tab:main} shows the result. Training further on \dataset{} raises the mean over the eight benchmarks from 55.4 to 57.9 and improves every one of them.

Training with GRPO, however, has a known limitation. GRPO takes its learning signal from the reward variance within a group of $G$ rollouts of one question, so a group whose rollouts all earn the same reward carries zero advantage and contributes nothing to the update. The literature calls this failure advantage collapse \citep{avspo2026, edgegrpo2025}, and we call such a group a zero-variance group. Prior work handles this limitation either by discarding the group and sampling new prompts until the batch is full \citep{yu2025dapo}, which spends more rollouts, or by reshaping the advantage of the group it already has \citep{rlzvp2025, ngrpo2025}.

Instead of drawing more groups, we change how each group is drawn: with $G = 8$ rollouts per question, we sample the first 4 rollouts normally. If these 4 rollouts are either all correct or all incorrect, we sample the last 4 rollouts under a top-token mask inside the reasoning span. Wherever the policy places more than $\tau = 0.95$ of its mass on one token, we drop that token and renormalize over the rest, so the model continues from a token that it samples rarely on its own. All 8 rollouts then enter the group advantage, so the intervention adds no rollouts. We call this \method{} (\methodshort{}). On top of \dataset{}, it raises the mean over the eight benchmarks from 57.9 to 59.3. Our model, \model{}, combines standard RL, \dataset{} and \methodshort{}. \ifpreprint We publicly release \else We will publicly release \fi the full dataset, the code that generates it and trains the model, and the \model{} checkpoint\ifpreprint\else\ shortly\fi, so that others can reproduce every result and build on them.

In summary, we make three contributions.
\begin{itemize}[leftmargin=*, itemsep=2pt, topsep=2pt]
\item \textbf{The \dataset{} framework.} We propose a framework that generates multi-hop questions for training video reasoning models (Section~\ref{sec:pipeline}).
\item \textbf{The \dataset{} dataset and benchmark.} \ifpreprint We release\else We will release\fi{} 22,550 training questions over 13,378 videos, together with 1,000 questions on 1,000 held-out videos as a benchmark built from the framework.
\item \textbf{\method{} (\methodshort{}).} We introduce a simple and lightweight method that recovers zero-variance groups, which improves the results of GRPO training on \dataset{}.
\end{itemize}

\ifpreprint\else
\begin{figure*}[t]
\centering
\includegraphics[width=0.9\linewidth]{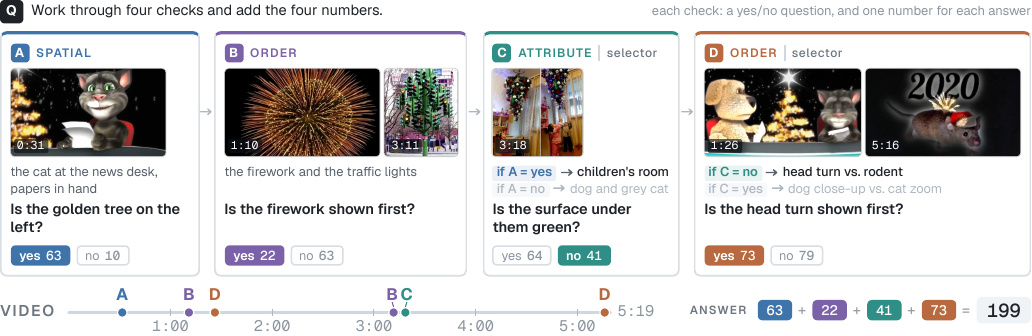}
\caption{\examplecaption}
\label{fig:example}
\end{figure*}
\fi

\label{sec:intro:end}

\begin{figure*}[t]
\centering
\includegraphics[width=0.72\linewidth]{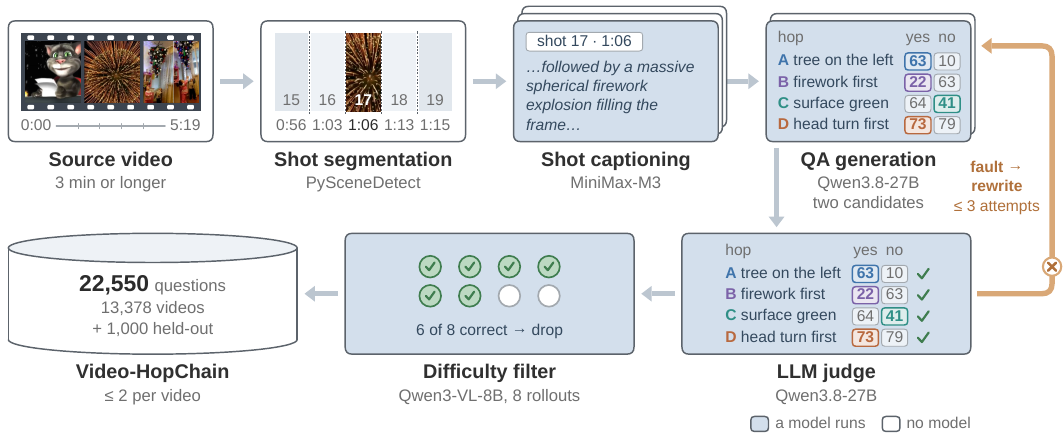}
\caption{The \dataset{} generator.}
\label{fig:pipeline}
\end{figure*}

\section{Related work}
\label{sec:related}

\paragraph{Training data for multimodal reinforcement learning.}
Reinforcement learning with verifiable rewards needs questions with one correct answer, so most multimodal datasets pool public benchmarks and instruction sets \citep{videor12025, onethinker2025}. However, audits find that a model answers a large share of long-video questions from text alone \citep{vidground2026}, shortcut-aware benchmarks measure when it answers from its priors \citep{mvp2025}, and probes show that an omnimodal model can encode a premise-perception mismatch that it never acts on \citep{sensesgap2026}. The usual response removes such shortcuts after the fact, with a text-only solver \citep{vidground2026}, inside a caption-grounded synthesis loop \citep{rewatch2025}, or by reducing text bias in the options \citep{textbias2026}. In contrast, we build every question around a multi-hop structure that ties every hop to visual evidence, which prior work applies to text \citep{musique2021, synthmultihop2026} and to video reasoning \citep{agqa2021, sung2026crit}, and which recent video benchmarks measure \citep{vrbench2025}. HopChain applies it to still images \citep{hopchain2026}, and we adapt it to video, where a hop becomes a yes/no question about a moment in time.

\paragraph{Reinforcement learning for video reasoning.}
Video-R1 applied GRPO to video with a temporal contrast term \citep{videor12025}, and later systems extend this recipe to spatio-temporal grounding \citep{videochatr1}, reasoning traces and thinking with video \citep{videorft, videothinker2025}, key-token attribution \citep{videoktr2026}, long video \citep{longvilar1}, and joint image-and-video training \citep{onethinker2025}. Almost all of them draw every group from the unmodified policy, and STRIVE changes only what a group is drawn over \citep{strive2026}. \methodshort{} instead changes the sampler itself, so it is orthogonal to these recipes.

\paragraph{Zero-variance groups.}
This shared sampler leaves in place the zero-variance limitation of Section~\ref{sec:twe}, which recent work calls advantage collapse \citep{avspo2026}. Objective-level variants reshape the estimator \citep{gspo2025, gvpo2025}, whereas other methods spend more compute, because they resample the group \citep{yu2025dapo}, reallocate rollouts across prompts \citep{reinforceada2025, knapsackrl2025}, branch a rollout \citep{treebranch2025}, or filter by difficulty during training \citep{odf2025}. A third family reuses the group it already has, by reshaping the advantage \citep{rlzvp2025, ngrpo2025} or by recovering signal from all-incorrect groups \citep{lens2025}, in some designs with a teacher model \citep{han2026rstg}. \methodshort{} also reuses the group, but it samples the second half of a zero-variance group differently, so that the group can regain variance.

\paragraph{Exploration and the loss at the token level.}
Our work also relates to methods that make the rollout more exploratory. Many reweight or restrict the gradient at high-entropy positions \citep{wang2025forking, cheng2025rwe} or at the tokens whose log-probability covaries most with the advantage \citep{cui2025entropymech}, while others act on low-probability tokens \citep{huang2025lpreg, lou2026taco}, on the probability mass after the rollout \citep{huo2026sps}, on noise in prompt or parameter space \citep{huang2026lope, psnrlvr2026}, or on the surprisal of the sampled token \citep{lv2026rsi, luo2026stare}. Closer to the sampler, FR3E explores from high-uncertainty decision points \citep{fr3e2025}, EDGE-GRPO corrects the rollout errors that drive the advantage to zero \citep{edgegrpo2025}, and CaSP acts in the loss on the top-1 probability that we use as our trigger \citep{peng2025casp}. The closest methods perturb the rollout as we do: EEPO regenerates the second half of a group after a transient weight update \citep{chen2025eepo}, and REFT resamples the first reasoning token from the top-$N$ candidates of the policy \citep{kim2026reft}. In contrast, \methodshort{} acts only on zero-variance groups, perturbs one token at a time in the sampler, and leaves the answer tokens untouched. Such a perturbation makes the rollout off-policy, as a mismatch between training and inference engines does \citep{tim2026}, which prior work handles by masking the discrepant tokens \citep{icepop2025} or by correcting their importance weight \citep{ctpo2026, impala2018}. For a deliberate perturbation \citep{noisyrollout}, prefix-guided methods remove the injected tokens from the loss \citep{adaprefix2026}, whereas LUFFY reshapes their gradient \citep{luffy2025}. We follow the importance-weight correction, because the sampling engine scores the forced token under the distribution it samples from, so the ratio of the clipped objective already weights that token (Section~\ref{sec:twe} and Appendix~\otherref{app:deriv}{I}\ofsupp{}). The tokens after a mask need no correction, since the unmodified policy samples them.

\section{\dataset{}}
\label{sec:data}
\ifpreprint\else
\begin{figure*}[t]
\centering
\includegraphics[width=0.72\linewidth]{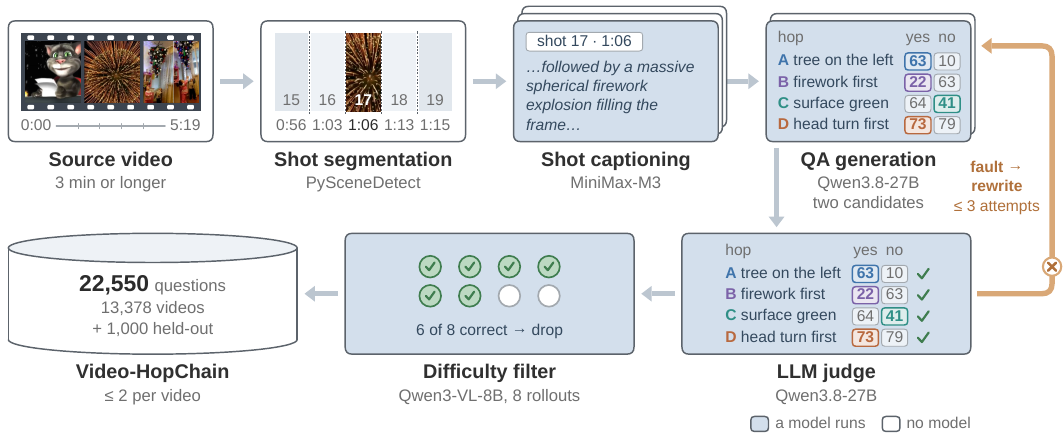}
\caption{The \dataset{} generator.}
\label{fig:pipeline}
\end{figure*}
\fi

\subsection{Question design}
\label{sec:design}

\dataset{} is a dataset of multi-hop video questions, in which every question chains $n$ hops over one video, with $n$ between three and six. Every hop carries two numbers, of which the first one counts when the answer to the hop is yes and the second one when the answer is no. The model then adds the numbers of all the hops, which makes this sum the answer to the question. Figure~\ref{fig:example} presents a sample of \dataset{} hop by hop.

For every hop we assign the two numbers by sampling integers at random in the range $1$ to $80$. We then resample them until no two combinations of answers give the same total. In addition, we keep the arithmetic a plain sum, so that the model only adds the numbers that its answers select. With this design, we intend the difficulty to come from the video, not the calculation.

A hop is a yes/no question that the video settles, and we use four categories of question for each hop, each over a different kind of visual evidence. The video settles three of these categories at one moment, whereas it settles the order category across two moments. We choose these four categories because a caption records this kind of information.

\begin{itemize}[leftmargin=*, itemsep=2pt, topsep=2pt]
\item An \emph{order} hop asks about the temporal order of two events, so it spans two moments.
\item A \emph{spatial} hop asks how two things are arranged inside one frame, such as left, right or middle, in terms anchored to the frame rather than to the body of the person shown.
\item An \emph{action} hop asks which physical action an agent performs, such as pour or lift.
\item An \emph{attribute} hop asks about a visible property of a named object, namely its color.
\end{itemize}

Beyond the category of each hop, we link the hops in one of two ways. In a \emph{flat} question, which we choose with probability 0.30, every hop names the moment it asks about and does not chain to another hop, so the model can answer the hops in any order. By contrast, in a \emph{selector} question, which we choose with probability 0.70, the answer to an earlier hop decides which of two moments a later hop asks about, so the model has to work through the hops in order.

\begin{figure*}[t]
\centering
\includegraphics[width=0.55\linewidth]{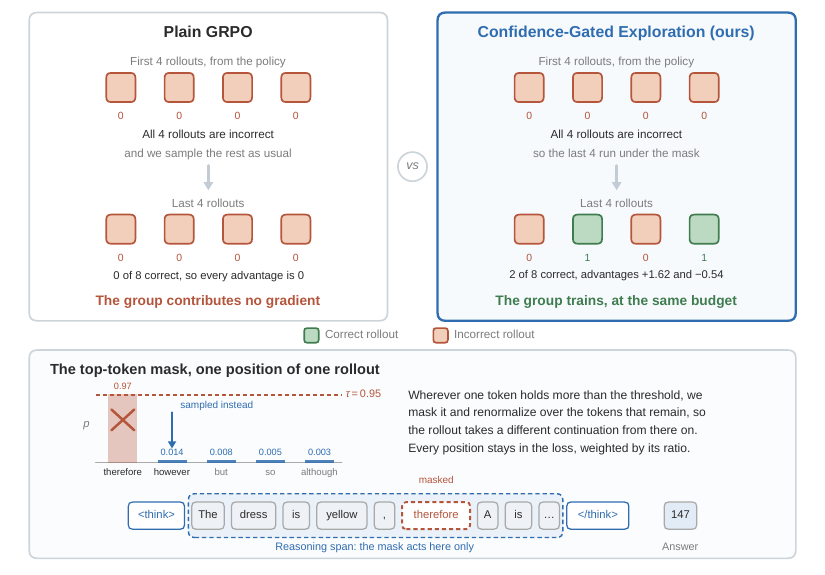}
\caption{\method{} on one group of $G$ rollouts.}
\label{fig:method}
\end{figure*}

\subsection{Generation pipeline}
\label{sec:pipeline}

Figure~\ref{fig:pipeline} shows the pipeline that turns raw video into questions. The source videos come from FineVideo \citep{finevideo2024}, from LongVILA \citep{longvila2024} as released with the OneThinker training data \citep{onethinker2025}, and from the long split of Vript \citep{vript2024}.

\paragraph{From video to captions.}
From these sources, we keep a source video only when it lasts at least three minutes, because we want the questions to stay hard and to force the model to reason over moments that lie far apart in time. We then cut the video into shots with PySceneDetect \citep{pyscenedetect}, and we caption every shot once with the MiniMax-M3 model \citep{lai2026minimax}.

\paragraph{Generating the question.}
Before we write a question from these captions, we fix its specification. To keep the questions diverse, we sample the hop count and the hop types with fixed probabilities. Table~\otherref{tab:corpus}{7}\ofsupp{} reports the hop counts of the corpus that these draws produced. This specification also fixes the way the hops link and the two numbers of each hop. We then use Qwen3.8-27B \citep{qwen38} as a question generator that writes two candidate questions per video from the captions and this specification.

\paragraph{Verifying and assembling.}
Once the generator writes the candidates, the same model re-examines every one of them against the same captions, and it passes a question only when it faults no hop. When the judge faults a hop, we send the question back to the generator, which writes it again from the same captions and the same specification. The judge then examines the new question. We repeat this loop until the judge reports no fault, but we drop the question when the judge still faults it after three attempts. The pipeline then ends with a difficulty filter, which removes every question that the base model already solves in at least 6 of 8 rollouts.

\subsection{Statistics and split}
\label{sec:stats}

Together, these stages yield 22,550 training questions over 13,378 videos and a held-out split of 1,000 questions over 1,000 further videos. We split by video, so no video appears on both sides. In addition, we draw the held-out side from the longer and more varied questions, stratified by hop count, video length and source.
\label{sec:data:end}

\section{\method{}}
\label{sec:method}
\subsection{Preliminaries}
\label{sec:prelim}

We train with Group Relative Policy Optimization (GRPO) \citep{grpo2024, deepseekr12025}. For a prompt $q$, which here is a video and a question, the policy $\pi_\theta$ samples a group of $G$ rollouts $\{o_i\}_{i=1}^{G}$, and a verifier then scores each one with a reward $r_i$. Reasoning work commonly rewards the answer and the response format together. We use this reward in this paper:
\begin{equation}
r_i = 0.8\,a_i + 0.2\,f_i ,
\label{eq:reward}
\end{equation}
where $a_i$ is $1$ when the answer matches the reference and $f_i$ is $1$ when the response follows the format that the system prompt asks for, namely a reasoning span inside the think tags and then an answer that carries the final integer in a boxed expression. Appendix~\otherref{app:corpus}{F}\ofsupp{} gives that system prompt in full. GRPO needs no value model, because the advantage $\hat{A}_i$ of a rollout is its reward standardized over the rewards of its own group. Every token of $o_i$ then carries that one value. With the importance ratio $\rho_{i,t}(\theta)$ between the current policy and the policy that sampled the rollout, we maximize the token-level clipped objective:
\begin{equation}
\begin{aligned}
\mathcal{J}(\theta) = \E_{q,\{o_i\}}\bigg[ &
\frac{1}{\sum_{i=1}^{G} |o_i|}
\sum_{i=1}^{G} \sum_{t=1}^{|o_i|} \\
& \min\!\Big(\rho_{i,t}(\theta)\,\hat{A}_i,\; \\
& \quad \clip\big(\rho_{i,t}(\theta),\, 1-\epsilon_{\text{low}},\, 1+\epsilon_{\text{high}}\big)\,\hat{A}_i\Big)
\bigg].
\end{aligned}
\label{eq:grpo}
\end{equation}
The clip range is asymmetric, following \citet{yu2025dapo}. Throughout, we use $G = 8$, $\epsilon_{\text{low}} = 0.2$, $\epsilon_{\text{high}} = 0.3$ and no KL penalty.

\subsection{\method{}}
\label{sec:twe}

The advantage of GRPO has a direct consequence. When every rollout of a group earns the same reward, the group has no reward variance, so every advantage is zero. We roll out and score such a group in full, and it then contributes no gradient. Following \citet{rlzvp2025}, we call such a group a \emph{zero-variance} group. These groups appear from both directions, because a question the model cannot solve returns all-incorrect rollouts, whereas a question it has learned returns all-correct ones. Among the methods that answer this problem, the closest design to ours is EEPO \citep{chen2025eepo}, which also regenerates part of a group after an intervention. \methodshort{} instead keeps the group and samples its second half differently, as Figure~\ref{fig:method} summarizes and as Algorithm~\otherref{alg:twe}{1}\ofsupp{} states.

\paragraph{When we encourage the model to explore.}
We draw each group in two waves of equal size, so the first wave takes $G_1 = 4$ of the 8 rollouts from $\pi_\theta$, and we then score them. If those 4 rollouts do not all earn the same accuracy, the group already carries variance, so we draw the second wave from $\pi_\theta$ unchanged. If they are all correct or all incorrect, however, the first wave is zero-variance, so we draw the second wave under the mask below instead. This check reads the accuracy $a_i$ of Equation~\ref{eq:reward} rather than the full reward $r_i$, so the format term $f_i$ alone never makes a group appear to carry variance. We intervene on both kinds of zero-variance group, all correct and all incorrect. To test this choice, we also ablate the alternative that intervenes on all-incorrect groups alone. This alternative is intuitive, because exploration appears most necessary on the questions that the model fails to solve, yet it performs worse, as Section~\ref{sec:ablations} reports.

\paragraph{Top-token mask.}
When the first wave is zero-variance, the second wave samples differently. Let $\ystar_t$ denote the token the policy finds most likely at position $t$ and $\pstar_t$ its probability. Wherever $\pstar_t$ exceeds $\tau$ and the position lies inside the reasoning span, between the think tags, we drop that token and share its probability over the rest:
\begin{equation}
\begin{aligned}
&\tilde{\pi}_\theta(\ystar_t) = 0,
\qquad
\tilde{\pi}_\theta(y) = \frac{\pi_\theta(y \mid q, o_{i,<t})}{1 - \pstar_t} \\
&\quad \text{for every other token } y .
\end{aligned}
\label{eq:mask}
\end{equation}
Here $\pi_\theta$ is the policy we train, $q$ is the prompt, $y$ is a candidate token at position $t$, and $o_{i,<t}$ is the part of rollout $i$ that the policy has already produced. Everywhere else the second wave samples from $\pi_\theta$ unchanged. Because we renormalize the remaining mass, the sampler draws from the policy's own alternatives in proportion to their probability, so the rollout continues from an alternative token that the policy samples rarely at that position. We set $\tau = 0.95$ (Section~\ref{sec:ablations} ablates $\tau$). We name the method after this gate, because the second wave explores exactly at the positions where the policy is most confident. Because the mask acts only inside the reasoning span, it never changes the answer tokens that the reward reads. The think tags that bound that span lie outside it, so the mask never removes them. Most token-level exploration methods trigger on the entropy or the surprisal of the next token, or on a statistic that also needs the advantage \citep{wang2025forking, lv2026rsi, luo2026stare, cui2025entropymech}. We follow this direction, but we use $\pstar_t$ as the trigger instead, because the mask acts on the top-1 token, so $\pstar_t$ measures exactly the confidence that the mask removes. A fixed top-1 probability still allows many different entropy values, so a trigger on $\pstar_t$ and a trigger on the entropy select different positions.

\paragraph{Forced tokens in the loss.}
We keep every position of every rollout in the sum of Equation~\ref{eq:grpo}, including the masked positions, which are the positions where the mask forces a token other than the top one. The advantage $\hat{A}_i$ still covers all $G$ rollouts of both waves together, so each rollout of the second wave contributes in the same way as a rollout of the first wave. A forced token is off-policy, and the importance ratio weights it without a special rule. Because the sampling engine scores the token under the masked and renormalized distribution, the ratio in Equation~\ref{eq:grpo} falls to $1 - \pstar_t \leq 0.05$ at the first update:
\begin{equation}
\rho_{i,t} = \frac{\pi_\theta(o_{i,t} \mid \cdot)}{\tilde{\pi}_\theta(o_{i,t} \mid \cdot)} = 1 - \pstar_t \;\leq\; 1 - \tau = 0.05 .
\label{eq:ratio}
\end{equation}
This value is far below the lower clip bound, so the clipped objective treats the forced token according to the sign of its advantage. For a positive advantage, it keeps the token at a small weight, so an alternative that leads to a better outcome gains a little probability. For a negative advantage, the clipped term is constant, so the token receives no gradient. If the second wave changes the outcome of a zero-variance group, every rollout in the group now receives a gradient. Otherwise, the group stays zero-variance and costs nothing more than before.

\paragraph{Cost.}
\methodshort{} adds only one barrier per prompt, since the second wave cannot start before we score the first wave. In our asynchronous trainer, however, we measure no significant change in throughput, as Appendix~\otherref{app:cost}{K}\ofsupp{} reports.
\label{sec:method:end}

\begin{table*}[!t]
\centering
\small
\setlength{\tabcolsep}{3.5pt}
\resizebox{0.90\linewidth}{!}{%
\begin{tabular}{lcccccccc|cc}
\toprule
model & VMME & PComp & VMMMU & Holmes & VCR & MMR-V & LVR & VRB & mean & in-domain \\
\midrule
Qwen3-VL-8B-Instruct & 64.0 & 28.1 & 63.5 & 40.7 & 32.0 & 42.7 & 72.6 & 74.7 & 52.3 & 13.4 \\
+ standard RL & 65.8 & 34.3 & 63.0 & 47.4 & 35.9 & 43.4 & 76.1 & 77.6 & 55.4 & 13.4 \\
+ standard RL + \methodshort{} & 67.9 & 34.4 & 62.5 & 48.5 & 35.1 & 46.4 & 78.8 & 79.5 & 56.6 & 16.4 \\
+ standard RL + \dataset{} & 68.6 & 36.2 & 64.7 & \textbf{48.6} & 42.1 & 45.9 & 77.8 & 79.4 & 57.9 & 21.2 \\
\ourrow + standard RL + \dataset{} + \methodshort{} & \textbf{69.2} & \textbf{37.4} & \textbf{67.5} & \textbf{48.6} & \textbf{44.9} & \textbf{46.6} & \textbf{78.9} & \textbf{81.0} & \textbf{59.3} & \textbf{23.2} \\
\bottomrule
\end{tabular}}

\caption{Main results with the best value of each column in bold. The shaded row is our model, \model{}.}
\label{tab:main}
\end{table*}
\begin{table*}[!t]
\centering
\small
\setlength{\tabcolsep}{4pt}
\renewcommand{\arraystretch}{0.92}
\resizebox{0.86\linewidth}{!}{%
\begin{tabular}{lcccccccc}
\toprule
model & \begin{tabular}[b]{@{}c@{}}Video-\\MME\end{tabular} & PerceptionComp & \begin{tabular}[b]{@{}c@{}}Video-\\MMMU\end{tabular} & \begin{tabular}[b]{@{}c@{}}Video-\\Holmes\end{tabular} & VCRBench & MMR-V & \begin{tabular}[b]{@{}c@{}}LongVideo-\\Reason\end{tabular} & VRBench \\
\midrule
\multicolumn{9}{l}{\emph{base model}} \\
Qwen3-VL-8B-Instruct$^\dagger$ & 64.0 & 28.1 & 63.5 & 40.7 & 32.0 & 42.7 & 72.6 & 74.7 \\
\midrule
\multicolumn{9}{l}{\emph{open-source video reasoning models}} \\
Video-R1 \citep{videor12025} & 61.4 & 26.3 & 52.4 & 36.5 & 48.0 & 36.3 & 68.1 & 69.5 \\
VideoChat-R1 \citep{videochatr1} & 60.0 & \underline{28.6} & 46.4 & 33.0 & \underline{48.2} & 36.1 & 67.2 & 61.5 \\
VideoRFT \citep{videorft} & 59.8 & -- & 51.1 & -- & -- & -- & -- & -- \\
Video-RTS \citep{videorts} & 63.0 & -- & 52.7 & 40.7 & -- & -- & -- & -- \\
Video-KTR \citep{videoktr2026} & 62.5 & -- & 53.1 & 42.7 & -- & -- & -- & -- \\
Video-Thinker \citep{videothinker2025} & -- & -- & -- & 43.2 & -- & -- & -- & \underline{80.7} \\
Video-o3 \citep{zeng2026videoo3} & \underline{66.5} & -- & 51.7 & 46.5 & -- & \underline{44.7} & -- & -- \\
Conan \citep{conan2026} & -- & -- & -- & 44.6 & \textbf{51.0} & 42.7 & 72.8 & \textbf{81.0} \\
LongVILA-R1 \citep{longvilar1} & 65.1 & -- & 51.0 & -- & -- & -- & 72.0 & -- \\
OneThinker \citep{onethinker2025} & \underline{66.5} & -- & \underline{66.2} & \textbf{48.7} & -- & -- & \textbf{79.2} & -- \\
\midrule
\ourrow \model{} (ours) & \textbf{69.2} & \textbf{37.4} & \textbf{67.5} & \underline{48.6} & 44.9 & \textbf{46.6} & \underline{78.9} & \textbf{81.0} \\
\bottomrule
\end{tabular}}

\caption{Comparison with open-source video reasoning models. A dash marks an unreported number, and the best value of each column is bold with the second best underlined. A dagger marks a result that we reproduce under our own setting.}
\label{tab:refrows}
\end{table*}

\begin{figure*}[t]
\centering
\includegraphics[width=0.8\linewidth]{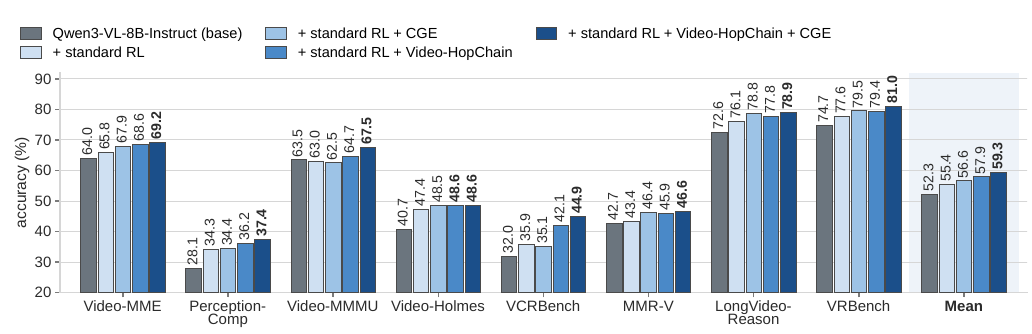}
\caption{Main result. Accuracy on the eight public video benchmarks and their mean, for the base model and for our four training runs. The darkest bars are our model, \model{}.}
\label{fig:mainresult}
\end{figure*}

\section{Experiments}
\label{sec:exp}

\subsection{Setup}
\label{sec:setup}

\paragraph{Training data.}
The \emph{general} dataset is a 105,993-row mixture of public multiple-choice video question answering data that prior video reasoning work assembled. Specifically, it draws on LLaVA-Video \citep{llavavideo2024} (72,421 rows), STAR \citep{star2021} (11,455), CLEVRER \citep{clevrer2020} (8,220), NExT-QA \citep{nextqa2021} (7,549) and PerceptionTest \citep{perceptiontest2023} (6,348). The \emph{multi-hop} dataset, in contrast, is \dataset{} with 22,550 rows. \ifpreprint We release \dataset{}\footnote{\releaseurl{\dataurl}}, our model \model{}\footnote{\releaseurl{\modelurl}}, the standard-RL checkpoint\footnote{\releaseurl{\baseurl}}, and the code that generates the dataset and trains the model\footnote{\releaseurl{\codeurl}}.\else The supplementary archive carries both \dataset{} and the code that generates it and trains our model.\fi

\paragraph{Model and training.}
We initialize the policy with Qwen3-VL-8B-Instruct \citep{qwen3vl2025}, and we train on 4 nodes of 8 H100 80GB GPUs, split as 2 rollout nodes and 2 trainer nodes, with a fully asynchronous GRPO trainer built on verl \citep{sheng2024hybridflow}. For the video input, we sample frames evenly over the whole video, namely 140 frames for \dataset{} and 24 frames for the general dataset. For \methodshort{}, we also use $\tau = 0.95$ and a first wave of $G_1 = 4$ rollouts.

\paragraph{Training stages.}
We run the following stages.
\begin{itemize}[leftmargin=*, itemsep=2pt, topsep=2pt]
\item \textbf{First stage, on the general dataset.} We train the base model on the general dataset. We then keep the checkpoint at which its accuracy peaks. This checkpoint is the warm start for the longer and more complicated reasoning traces that \dataset{} asks for. We call this run \emph{standard RL}.
\item \textbf{Second stage, on \dataset{} with plain GRPO.} We start this run from the single checkpoint of the first stage, and we then train on \dataset{} with plain GRPO.
\item \textbf{Second stage, on \dataset{} with \methodshort{}.} We start this run from the same checkpoint, but we train on \dataset{} with \methodshort{} instead of plain GRPO. This run gives our model, \model{}.
\item \textbf{\methodshort{} on the general dataset.} We start this run from the base model and enable the method from the first step.
\end{itemize}

\paragraph{Benchmarks.}
To measure these runs, we report eight public video benchmarks, namely Video-MME \citep{videomme2024}, PerceptionComp \citep{perceptioncomp2026}, Video-MMMU \citep{videommmu2025}, Video-Holmes \citep{videoholmes2025}, VCRBench \citep{vcrbench2025}, MMR-V \citep{mmrv2025}, LongVideo-Reason \citep{longvilar1} and VRBench \citep{vrbench2025}. We report VCRBench on its multiple-choice subset. We evaluate every benchmark with the lmms-eval framework \citep{lmmseval2024}, under the same settings for every run, and we give the model the same system prompt that it sees during training. For every benchmark, we sample 100 frames evenly over the whole video, and we decode with at most 501,760 pixels per frame, a context of 33,792 tokens and at most 16,384 generated tokens. We report the accuracy on the 1,000 held-out \dataset{} questions.

\begin{figure*}[t]
\centering
\includegraphics[width=0.70\linewidth]{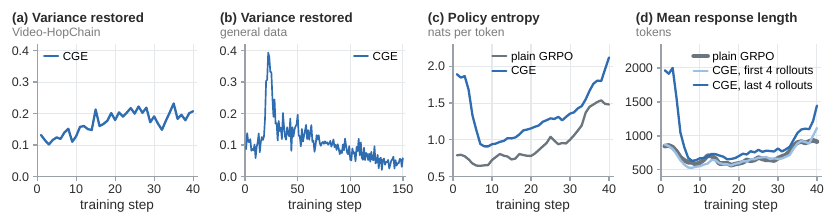}
\caption{Four training metrics of our runs, over the training steps.}
\label{fig:signal}
\end{figure*}

\subsection{Main results}
\label{sec:main}

\textbf{The dataset improves the model, and the method improves it further.}
On these eight public benchmarks (Table~\ref{tab:main}\ifpreprint\ and Figure~\ref{fig:mainresult}\fi), a second stage on \dataset{} raises the mean and improves every one of them. Enabling \methodshort{} on top of that dataset raises the mean again. In addition, our model holds the best value of every column, where it ties the \dataset{} run on Video-Holmes.

\textbf{Our model outperforms open-source video reasoning models.}
Table~\ref{tab:refrows} places our model beside ten open-source video reasoning models on the same eight benchmarks. It gives the best value on most of these benchmarks, ties with Conan on VRBench, and comes within 0.3 points of OneThinker on Video-Holmes and on LongVideo-Reason. We compare released systems rather than recipes, because the ten models differ in base model, training data and compute.

\subsection{Ablations}
\label{sec:ablations}

\begin{table}[t]
\centering
\small
\setlength{\tabcolsep}{5pt}
\begin{tabular}{lcc}
\toprule
run & $\tau$ & best mean \\
\midrule
plain GRPO & --- & 66.8 \\
\methodshort{} & 0.80 & 65.0$^\dagger$ \\
\methodshort{} & 0.85 & 64.9$^\dagger$ \\
\methodshort{} & 0.90 & \textbf{72.6} \\
\methodshort{} & 0.95 & 72.5 \\
\midrule
\methodshort{}, all-incorrect only & 0.95 & 66.7 \\
\bottomrule
\end{tabular}
\caption{Ablations on OpenMMReasoner, as the best mean over the six image benchmarks of OpenMMReasoner. The last row intervenes on all-incorrect groups alone, whereas every other \methodshort{} run intervenes on all-correct groups as well. A dagger marks a run that never learns the answer format.}
\label{tab:ablation}
\end{table}

We ablate the two choices of Section~\ref{sec:twe}, namely the threshold $\tau$ and the groups we intervene on. We run every one of them on the OpenMMReasoner RL dataset \citep{openmmreasoner2025}, a 74K-sample image dataset. \methodshort{} makes no assumption about the modality of the prompt and therefore applies to an image dataset as it does to video, and every run is a cold start from the same base model, with the same reward and the same compute budget as the video runs. We score every run on the six public image benchmarks that OpenMMReasoner evaluates on, namely CharXiv \citep{charxiv2024}, DynaMath \citep{dynamath2024}, MathVerse \citep{mathverse2024}, MathVista \citep{mathvista2024}, MMMU \citep{mmmu2024} and WeMath \citep{wemath2024}, and Table~\ref{tab:ablation} reports the best mean over those six benchmarks.

\textbf{The threshold needs no tuning above 0.90.}
At $\tau = 0.90$ and $\tau = 0.95$, \methodshort{} reaches 72.6 and 72.5, which is more than five points above plain GRPO at the same compute budget, and the two thresholds stay within 0.1 points of each other. At $\tau = 0.85$ and $\tau = 0.80$, by contrast, the model never learns to emit the answer format, the response length grows from 380 to over 13,000 tokens, and the fraction of clipped tokens rises from zero to 0.69. We read these two failures as one mechanism, because a lower threshold makes the mask eligible at positions of ordinary uncertainty rather than only where the policy is nearly certain, so the second wave fires at many more positions of the same rollout and the reasoning span keeps growing instead of closing. We therefore apply the mask only in the near-certain regime, under which the reasoning still converges.

\textbf{Both kinds of zero-variance group need the intervention.}
A natural alternative to Section~\ref{sec:twe} is to intervene only when the first wave is all incorrect, because a group that the model already solves appears to need no exploration. However, most zero-variance groups on this dataset are all correct, so this alternative lowers the share of intervened groups from 0.79 to 0.29, and the run then peaks at 66.7, which is 5.8 points below \methodshort{} at the same threshold and no better than plain GRPO. For this reason we intervene on all-correct and all-incorrect groups alike, and Appendix~\otherref{app:ablations}{D}\ofsupp{} reports both ablations in full.

\label{sec:exp:end}

\section{Analysis}
\label{sec:analysis}

\textbf{\methodshort{} recovers one zero-variance group in six.}
The first panel of Figure~\ref{fig:signal} tracks the share of the intervened groups whose second wave restores the variance of the group. This share grows over training, from about one group in seven to one in five. Almost every intervened group is a question the model cannot yet solve, because on \dataset{} the first wave comes back all incorrect far more often than all correct. However, the mask changes the outcome of an all-correct group much more often than that of an all-incorrect one, so we attribute the rise to the growing share of all-correct groups. The restored variance also comes from new solutions rather than from noise, because the second wave solves questions that the first wave never solves, and it answers as accurately as the first wave does. When we count both waves, therefore, \methodshort{} gives about half again as many groups with a gradient as the first wave alone. Appendix~\otherref{app:signal}{G}\ofsupp{} gives the full numbers.

\textbf{\methodshort{} holds the policy entropy above plain GRPO.}
While the first panel counts groups, the third panel tracks the token entropy of the policy on its own rollouts, where \methodshort{} stays 0.2 to 1.1 nats above plain GRPO for the whole run, although both runs reach the same training reward. Such a margin can matter, because reinforcement learning for reasoning models tends to lose entropy as the reward rises, until the policy stops exploring and the reward saturates, a failure that prior work calls entropy collapse \citep{cui2025entropymech}. \methodshort{} keeps more room to explore. We attribute this margin to the masked positions and to the tokens that follow them, because one alternative token carries a rollout along a path the policy samples rarely.

\textbf{The dataset and the method fit together.}
Because the mask acts only inside the reasoning span, the number of positions it can change grows with the length of that span, and a multi-hop question of \dataset{} therefore offers it many more positions than a multiple-choice question does. On the general dataset, the second wave restores fewer intervened groups on average than on \dataset{}.

\label{sec:analysis:end}

\section{Conclusion}
\label{sec:conclusion}

We presented \dataset{}, a dataset of multi-hop questions over videos. A second stage of training on this dataset improves general video understanding and reasoning beyond what standard RL on general video data reaches. Because such training also produces zero-variance groups, we then introduced \method{}, which resamples the second half of a zero-variance group under a top-token mask. The method therefore recovers part of these groups without additional rollouts, and it improves the results further. \ifpreprint We publicly release \else We will publicly release \fi the dataset, the data generation code, the training code and \model{}\ifpreprint\else\ shortly\fi, and we hope that both contributions help future work on video reasoning.%
\label{sec:conclusion:end}%

\label{sec:endbody}

{
    \small
    \bibliographystyle{ieeenat_fullname}
    \bibliography{refs}
}

\ifwithsupp
\ifpreprint\else
\clearpage
\setcounter{page}{1}
\maketitlesupplementary
\fi
\appendix

\section*{Appendix contents}

\begin{itemize}[leftmargin=*, itemsep=3pt, topsep=3pt]
\item \textbf{Appendix~\ref{app:alg}, Algorithm.} \methodshort{} for one prompt, step by step.
\item \textbf{Appendix~\ref{app:limits}, Limitations.} What this study does not show.
\item \textbf{Appendix~\ref{app:config}, Configuration.} Every hyperparameter of the two runs.
\item \textbf{Appendix~\ref{app:ablations}, Ablations.} The threshold $\tau$ and the groups we intervene on, measured on image data.
\item \textbf{Appendix~\ref{app:imagetransfer}, Image benchmark results.} The effect of this training on general image ability.
\item \textbf{Appendix~\ref{app:corpus}, Dataset details.} Statistics, row format, system prompt and the held-out split.
\item \textbf{Appendix~\ref{app:signal}, Group-level analysis of the second wave.} Which groups the second wave returns a gradient to.
\item \textbf{Appendix~\ref{app:drops}, Token-level analysis of the mask.} Where the mask acts and what it removes.
\item \textbf{Appendix~\ref{app:deriv}, Derivations.} The entropy bounds and the importance ratio at a masked position.
\item \textbf{Appendix~\ref{app:update}, Effect of the second wave on the update.} What the second wave changes in the update.
\item \textbf{Appendix~\ref{app:cost}, Computational cost and response length.} The cost of the second wave, and its effect on response length.
\end{itemize}

\newpage
\section{Algorithm}
\label{app:alg}

\begin{algorithm}[H]
\small
\caption{\method{} for one prompt}
\label{alg:twe}
\begin{algorithmic}[1]
\Require prompt $q$, policy $\pi_\theta$, group size $G$, threshold $\tau$
\State sample $o_1, \dots, o_{G/2} \sim \pi_\theta(\cdot \mid q)$ and score $r_1, \dots, r_{G/2}$ with accuracies $a_1, \dots, a_{G/2}$ \Comment{first wave}
\State $g \gets 1$ if $a_1 = \cdots = a_{G/2}$, else $0$ \Comment{is the first wave all correct or all incorrect?}
\For{$i = G/2 + 1, \dots, G$} \Comment{second wave}
    \For{$t = 1, 2, \dots$ until end of response}
        \State $\pstar_t \gets \max_y \pi_\theta(y \mid q, o_{i,<t})$
        \If{$g = 1$ and $\pstar_t > \tau$ and $t$ is inside the reasoning span}
            \State sample $o_{i,t} \sim \tilde{\pi}_\theta(\cdot \mid q, o_{i,<t})$ \Comment{top-token mask, Eq.~\otherref{eq:mask}{3}}
        \Else
            \State sample $o_{i,t} \sim \pi_\theta(\cdot \mid q, o_{i,<t})$
        \EndIf
    \EndFor
    \State score $r_i$
\EndFor
\State compute $\hat{A}_1, \dots, \hat{A}_G$ over all $G$ responses \Comment{standardized within the group}
\State \Return $\{(o_i, \hat{A}_i)\}_{i=1}^{G}$ \Comment{every position enters the loss}
\end{algorithmic}
\end{algorithm}

\section{Limitations}
\label{app:limits}

We train one base model at one scale, so we do not test whether our results hold at other scales. In addition, we generate and verify \dataset{} automatically, and although this dataset improves both the in-domain measure and the public benchmarks, a manually labeled and manually verified dataset could give different results. Finally, we do not run a few ablations, namely the first-wave size $G_1$, the restriction of the mask to the reasoning span, and an adaptive $\tau$.

\section{Configuration}
\label{app:config}

Table~\ref{tab:config} lists the settings of the two video runs, read from their resolved configurations. In this table, the plain GRPO column reports the plain GRPO run, whereas the \methodshort{} column reports the run that enables the mask, and these two runs differ only in the exploration settings.

\begin{table*}[t]
\centering\small
\setlength{\tabcolsep}{4pt}
\begin{tabular}{p{6.5cm}p{2.5cm}p{3.9cm}}
\toprule
setting & plain GRPO & \methodshort{} \\
\midrule
\multicolumn{3}{l}{\emph{data}} \\
maximum prompt length & 5,376 & 5,376 \\
maximum response length & 16,384 & 16,384 \\
\midrule
\multicolumn{3}{l}{\emph{optimizer}} \\
learning rate & $1 \times 10^{-6}$ & $1 \times 10^{-6}$ \\
warmup steps & 25 & 25 \\
weight decay & 0.1 & 0.1 \\
PPO mini-batch size & 16 & 16 \\
micro-batch size per GPU & dynamic & dynamic \\
maximum tokens per actor micro-batch & 26,004 & 26,004 \\
PPO epochs per batch & 1 & 1 \\
\midrule
\multicolumn{3}{l}{\emph{objective}} \\
advantage estimator & GRPO & GRPO \\
advantage normalized by group std. & yes & yes \\
clip range, low (Eq.~\otherref{eq:grpo}{2}) & 0.2 & 0.2 \\
clip range, high & 0.3 & 0.3 \\
dual-clip constant & 10 & 10 \\
KL loss & off & off \\
KL coefficient & 0 & 0 \\
KL term in the reward & off & off \\
entropy coefficient & 0 & 0 \\
loss aggregation & token-mean & token-mean \\
old log-probabilities (Eq.~\otherref{eq:ratio}{4}) & from the engine & from the engine \\
extra cross-engine correction & none & none \\
\midrule
\multicolumn{3}{l}{\emph{rollout}} \\
frames per video & 140 & 140 \\
maximum pixels per frame & 50,176 & 50,176 \\
group size $G$ & 8 & 8 \\
sampling temperature & 1.0 & 1.0 \\
top-$p$ & 1.0 & 1.0 \\
top-$k$ & $-1$, off & $-1$, off \\
tensor parallel size & 2 & 2 \\
GPU memory fraction & 0.8 & 0.8 \\
\midrule
\multicolumn{3}{l}{\emph{exploration}} \\
\methodshort{} enabled & no & yes \\
first-wave fraction & --- & 0.5, $G_1 = 4$ \\
gate: first-wave variance at most & --- & 0 \\
reward the check uses & --- & accuracy \\
intervene on all-correct groups too & --- & yes \\
trigger & --- & $p^\star_t > \tau$ \\
mask threshold $\tau$ (Eq.~\otherref{eq:mask}{3}) & --- & 0.95 \\
tokens masked per position & --- & 1 \\
positions eligible & --- & reasoning span \\
cap on masks per response & --- & 16,384 \\
forced tokens kept in the loss & --- & yes \\
\midrule
\multicolumn{3}{l}{\emph{schedule and hardware}} \\
epochs over the dataset & 2 & 2 \\
trainer nodes $\times$ GPUs & $2 \times 8$ & $2 \times 8$ \\
rollout nodes $\times$ GPUs & $2 \times 8$ & $2 \times 8$ \\
staleness threshold & 0.5 & 0.5 \\
parameter-sync interval & 4 steps & 4 steps \\
\bottomrule
\end{tabular}

\caption{Configuration of the plain GRPO run and the \methodshort{} run.}
\label{tab:config}
\end{table*}

\section{Ablations}
\label{app:ablations}

\paragraph{Scope of the ablations.}
We ablate two choices of Section~\otherref{sec:twe}{4.2}, namely the threshold $\tau$ and the groups we intervene on. We run every one of them on the OpenMMReasoner RL dataset \citep{openmmreasoner2025}, a 74K-sample image dataset, and we hold everything else fixed, namely the same base model, the same algorithm, the same reward and the same compute budget as the video runs. We run them on image data because an image run logs many more validation points than a video run of the same wall-clock budget, so it locates the best accuracy of a run far more precisely. We therefore treat these ablations as evidence about \methodshort{} itself rather than as measurements of video accuracy.

\paragraph{Setting.}
On that image dataset, every run here measures \methodshort{} alone, because \methodshort{} makes no
assumption about the modality of the prompt and therefore applies to an image dataset as it does to
video. Every run is a cold start from the base model on the same data, and we then score it on the six
public image benchmarks that OpenMMReasoner evaluates on, namely CharXiv \citep{charxiv2024},
DynaMath \citep{dynamath2024}, MathVerse \citep{mathverse2024}, MathVista \citep{mathvista2024},
MMMU \citep{mmmu2024} and WeMath \citep{wemath2024}. Every mean we report on this dataset is the
mean over those six benchmarks.

\paragraph{Threshold.}
Figure~\ref{fig:taupeaks} gives the best mean accuracy of the threshold sweep, together with plain
GRPO as the run without any exploration. At $0.90$ and $0.95$ both runs learn the answer format and reach more than
five points above plain GRPO at the same compute budget. Moreover, the two thresholds stay within 0.1
points of each other, so the threshold needs no tuning inside that range. At $0.85$ and $0.80$, by
contrast, the model never learns to emit the answer format, the response length grows from 380 to over
13,000 tokens, the fraction of clipped tokens rises from zero to 0.69, and the step time grows
thirteenfold. The accuracy at $\tau = 0.85$ rises again late in that run, however, and we attribute
this rise to length inflation rather than to recovery.

We read these two failures as one mechanism. A lower threshold makes the mask eligible at positions
of ordinary uncertainty, not only where the policy is nearly certain, so the second wave fires at
many more positions of the same rollout. The model then never settles on a line of reasoning,
because at every step where it starts to commit, the mask moves it off the token it was about to
write. The reasoning span therefore keeps growing instead of closing, which is what the rise from
380 to over 13,000 tokens measures. The response finally meets the token budget rather than an end
of its own, and the fraction of clipped tokens of 0.69 is that collision. A clipped response stops
inside the reasoning span, so it never reaches the answer that the format asks for, and the format
rate falls with it. Moreover, the mask applies only inside
the reasoning span, between the think tags, as Section~\otherref{sec:twe}{4.2} defines it, so a token of the
answer block is never eligible for the mask at any threshold. The format degrades because the
reasoning never ends, not because the mask deletes the format. We therefore apply the mask only
where the policy is nearly certain, since that is the setting under which the reasoning still
converges.

\begin{figure*}[t]
\centering
\includegraphics[width=0.52\linewidth]{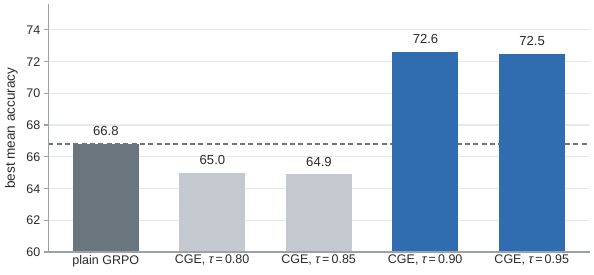}
\caption{Best mean accuracy over the six image benchmarks of OpenMMReasoner, for plain GRPO and for
\methodshort{} at four thresholds. The dashed line marks plain GRPO, and the two gray bars are the
runs that never learn the answer format.}
\label{fig:taupeaks}
\end{figure*}

\paragraph{Choice of the intervened groups.}
In Section~\otherref{sec:twe}{4.2}, we intervene on all-correct and all-incorrect groups alike. A natural alternative is to intervene only when the first wave is all incorrect, because a group that the model already solves appears to need no exploration. However, Table~\ref{tab:trigger} shows that this alternative fails. Because most zero-variance groups are all correct, a rule that skips them reduces the share of intervened groups from 0.79 to 0.29, and the run then peaks at 66.7, which is 5.8 points below \methodshort{} and no better than plain GRPO, before it collapses. For this reason we intervene on all-correct and all-incorrect groups alike.

\begin{table*}[t]
\centering
\footnotesize
\setlength{\tabcolsep}{4pt}
\begin{tabular}{llcc}
\toprule
we intervene on & $\tau$ & best mean accuracy (step) & vs.\ \methodshort{} \\
\midrule
all-correct and all-incorrect & 0.95 & 72.5 (180) & --- \\
all-incorrect only & 0.95 & 66.7 (50) & $-5.8$ \\
\bottomrule
\end{tabular}
\caption{Effect of the choice of intervened groups, measured on OpenMMReasoner.}
\label{tab:trigger}
\end{table*}

\section{Image benchmark results}
\label{app:imagetransfer}

Because every training stage of this paper uses video, we check whether this training lowers general
image ability. We therefore evaluate the same checkpoints of Table~\otherref{tab:main}{1} on six public
image benchmarks, namely AI2D \citep{ai2d2016}, MathVista \citep{mathvista2024}, MMBench-EN
\citep{mmbench2024}, MMMU \citep{mmmu2024}, MMStar \citep{mmstar2024} and RealWorldQA
\citep{realworldqa2024}, for a total of 11,527 questions. We take the multiple-choice split of
AI2D, the testmini split of MathVista, the English dev split of MMBench and the validation split of
MMMU, and on MMMU we keep the multiple-choice questions only. We score every benchmark by exact match on
a multiple-choice letter or on a numeric answer, so we use no model as a judge. Because we otherwise
follow the settings of Section~\otherref{sec:setup}{5.1}, we measure these image numbers in the same way as
the video numbers of Table~\otherref{tab:main}{1}. Table~\ref{tab:imagetransfer} then reports the result.

\begin{table*}[t]
\centering
\footnotesize
\setlength{\tabcolsep}{3.5pt}
\begin{tabular}{lcccccc|c}
\toprule
model & AI2D & MathVista & MMBench & MMMU & MMStar & RealWorldQA & mean \\
 & \scriptsize 3{,}088 & \scriptsize 998 & \scriptsize 4{,}329 & \scriptsize 847 & \scriptsize 1{,}500 & \scriptsize 765 & \\
\midrule
Qwen3-VL-8B-Instruct & 79.5 & \textbf{75.2} & 88.3 & 61.6 & \underline{67.5} & 63.3 & 72.6 \\
+ standard RL & 82.4 & 71.2 & 89.3 & 62.3 & \textbf{67.7} & 65.0 & 73.0 \\
+ standard RL + \methodshort{} & \textbf{83.3} & 70.5 & \textbf{89.6} & \underline{62.8} & 66.3 & \textbf{69.8} & \underline{73.7} \\
+ standard RL + \dataset{} & 82.7 & 71.6 & \underline{89.4} & 62.2 & 66.8 & 66.3 & 73.2 \\
\ourrow + standard RL + \dataset{} + \methodshort{} & \underline{82.8} & \underline{73.0} & 89.2 & \textbf{62.9} & \underline{67.5} & \underline{67.3} & \textbf{73.8} \\
\bottomrule
\end{tabular}
\caption{Image understanding, accuracy in percent, with the number of questions under each
benchmark, the best value of each column in bold and the second best underlined. The shaded row is our model, \model{}.}
\label{tab:imagetransfer}
\end{table*}

\section{Dataset details}
\label{app:corpus}

This appendix gives the statistics of the dataset, the row format of the released dataset, the system prompt, and how we draw the held-out split.

\paragraph{Dataset statistics.}
Table~\ref{tab:corpus} reports the size of the two splits, the hop counts and the range of the ground-truth answer, which is the integer that the hops of a question sum to.

\begin{table*}[t]
\centering
\footnotesize
\setlength{\tabcolsep}{4pt}
\begin{tabular}{lrr}
\toprule
 & train & held-out \\
\midrule
questions & 22,550 & 1,000 \\
videos & 13,378 & 1,000 \\
questions per video & 1.69 & 1.00 \\
questions with 3 / 4 / 5 / 6 hops & 595 / 12,604 / 7,527 / 1,824 & 0 / 302 / 450 / 248 \\
hops by type: order / spatial / action / attribute & 44,416 / 22,476 / 21,372 / 12,516 & 2,082 / 1,141 / 1,072 / 651 \\
ground-truth answer, smallest to largest & 20 to 436 & 52 to 366 \\
ground-truth answer, median & 178 & 198 \\
\bottomrule
\end{tabular}
\caption{\dataset{} statistics for the training split and the held-out split.}
\label{tab:corpus}
\end{table*}

\paragraph{Row format.}
Each training row of \dataset{} holds a two-message prompt, namely the system prompt of the dataset, which the released code carries, and a user turn that starts with the video placeholder and continues with the question text. The row also carries the video path with its decode contract, the ground-truth answer as a string, and an extra-information record with the answer, the hop count, the hop types, and the question and video identifiers. We store the hop records, with their moment references, mappings, values and caption quotes, beside the dataset for auditing, but we keep them out of the training rows.

\paragraph{System prompt.}
Every training row and every held-out row carries this system prompt, so we evaluate every model we train with the same prompt.

\nolinenumbers
\begin{lstlisting}
You are a careful reasoning assistant. ALWAYS respond in this EXACT format:

<think>step-by-step reasoning</think>
<answer>\boxed{final_answer}</answer>

Examples:

Q: 7 x 8?
<think>7 x 8 = 56.</think>
<answer>\boxed{56}</answer>

Q: A right triangle has legs of length 3 and 4. What is the hypotenuse?
<think>By the Pythagorean theorem, c^2 = 3^2 + 4^2 = 9 + 16 = 25, so c = 5.</think>
<answer>\boxed{5}</answer>

For multiple-choice, put the letter, e.g. \boxed{B}.
Always wrap reasoning in <think>...</think> and answer in <answer>\boxed{...}</answer>. No text outside these tags.
\end{lstlisting}
\linenumbers

\paragraph{The question of Figure~\otherref{fig:example}{2}.}
The dataset row of this question, held-out question \texttt{2Wa\_OoZyEk8::1}, carries the full text below, without the video placeholder.

\nolinenumbers
\begin{lstlisting}
Work through five checks and add the five numbers.

First, in the scene where a cat sits at a circular table wearing a red harness with silver buckles, look at the position of the golden Christmas tree: if the tree is on the left side of the frame, let A be 63; otherwise let A be 10.

Second, compare two moments: the moment a spherical firework explosion fills the frame, and the moment an outdoor scene shows a tall green pole supporting a cluster of traffic lights. If the firework explosion is shown before the traffic light scene, let B be 22; otherwise let B be 63.

Third, use the first check to choose where to look: if the tree is on the left side of the frame, look at the room where two children stand beneath an upside-down Christmas tree; otherwise look at the scene where a dog and a grey cat rest on a flat surface with the cat holding one paw raised near its face. In whichever of those two places you were sent to, look at the surface beneath the characters: if the surface is green, let C be 64; otherwise let C be 41.

Fourth, use the third check to choose which pair of moments to compare: if the surface is green, compare the moment the dog's face fills the frame in close-up with out-of-focus colored lights on the left against the moment the camera zooms in until the grey cat's face fills most of the frame; if the surface is not green, compare the moment the dog's head is turned toward the cat with a teal banner at the bottom of the frame against the moment a small rodent in a Santa hat is shown with fireworks bursting around it. In whichever pair you were sent to, if the first-named moment is shown before the second-named moment, let D be 73; otherwise let D be 79.

Fifth, in the news desk scene where red illuminated elements are visible below the desk's cyan rim, look at what the dog is doing: if the dog is turning its head toward the other character, let E be 70; if the dog is looking down at its paper, let E be 42.

Add A, B, C, D, and E.
\end{lstlisting}
\linenumbers

\paragraph{Held-out split.}
We split the dataset by video with seed 42, and we then draw the held-out split with seed 1234 from the questions with at least four hops and at least two distinct hop types, with exactly one question per video, and we stratify this split on three axes at once. By hop count it holds 302, 450 and 248 questions with 4, 5 and 6 hops. By video length it holds 151 questions on videos under 40 segments, 301 on 40 to 70 segments, 300 on 70 to 110 segments, and 248 on longer videos. By source it holds 467 questions from LongVILA, 358 from FineVideo and 175 from Vript.

\section{Group-level analysis of the second wave}
\label{app:signal}

Table~\ref{tab:signal} gives the numbers behind the observations of Section~\otherref{sec:analysis}{6}, and we
measure every row on the run of our model in Table~\otherref{tab:main}{1}, over the same window that
Figure~\otherref{fig:signal}{5} plots. We report the mean over the first half of that window, which we label early,
and the mean over the second half, which we label later.

\begin{table*}[t]
\centering
\footnotesize
\setlength{\tabcolsep}{5pt}
\begin{tabular}{lcc}
\toprule
share of & early & later \\
\midrule
\multicolumn{3}{l}{\emph{what the first wave produces, over all groups}} \\
groups whose 4 rollouts are all incorrect & 0.73 & 0.64 \\
groups whose 4 rollouts are all correct & 0.03 & 0.08 \\
groups that already carry variance & 0.24 & 0.28 \\
\midrule
\multicolumn{3}{l}{\emph{what the second wave adds, over the intervened groups}} \\
groups whose variance the second wave restores & 0.15 & 0.20 \\
all-correct groups whose outcome the mask changes & 0.70 & 0.61 \\
all-incorrect groups in which the mask produces a correct rollout & 0.13 & 0.15 \\
groups in which the second wave solves what the first wave never solves & 0.12 & 0.13 \\
accuracy of the second wave minus the first & $+0.02$ & $+0.01$ \\
\midrule
\multicolumn{3}{l}{\emph{what the group contributes to the update}} \\
groups that carry a gradient, first wave alone & 0.24 & 0.28 \\
groups that carry a gradient, both waves together & 0.36 & 0.42 \\
questions solved by at least one of the 8 rollouts & 0.37 & 0.45 \\
questions solved by at least one of the first 4 & 0.28 & 0.36 \\
\bottomrule
\end{tabular}
\caption{The signal that the second wave recovers, measured on \dataset{}.}
\label{tab:signal}
\end{table*}

\begin{figure*}[t]
\centering
\includegraphics[width=\linewidth]{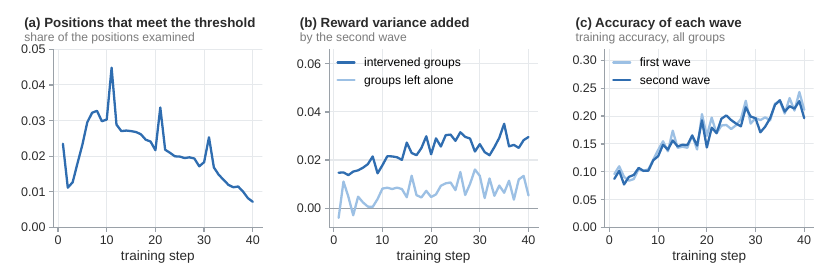}
\caption{Three measurements of the second wave on \dataset{}, over the training steps. The left panel gives the share of the positions the mask examines whose top-1 probability exceeds $\tau$, whereas the middle panel gives the reward variance the second wave adds to a group, both on the intervened groups and on the groups the mask leaves alone, which already carry variance in their first wave. The right panel then gives the training accuracy of each wave over all groups.}
\label{fig:maskact}
\end{figure*}

\paragraph{The variance the second wave adds.}
Table~\ref{tab:signal} counts the groups whose variance the second wave restores, and we also measure how much variance it adds. For a group of $G$ rollouts whose first wave answers a share $p$ of them correctly, we take as reference the accuracy variance that $G$ independent rollouts with success rate $p$ give on average, $p(1-p)(G-1)/G$, and we subtract it from the accuracy variance we measure after the second wave and report the difference, which the middle panel of Figure~\ref{fig:maskact} plots. On an intervened group the first wave is all correct or all incorrect, hence $p$ is 0 or 1, the quantity we subtract is exactly zero, and the whole of the remaining variance follows from the mask. This difference grows from 0.020 early in the run to 0.028 later, whereas on the groups that the mask leaves alone the same difference reaches only 0.005 early and 0.009 later, so the subtraction leaves almost no variance on those groups.

\paragraph{The accuracy of each wave.}
The right panel of Figure~\ref{fig:maskact} tracks the accuracy of each wave. Over all groups the two waves reach the same accuracy for the whole run, at 0.17 for the first wave and 0.16 for the second, whereas over the intervened groups alone the second wave is more accurate than the first by 0.02 early and by 0.01 later, as the accuracy row of Table~\ref{tab:signal} reports. We measure the same on the general dataset, where the first wave reaches 0.81 and the second reaches 0.80 over the whole run. Therefore the second wave answers as accurately as the first, and it answers more accurately on the groups the mask acts on, while it follows a path that the policy would otherwise not have taken.

\paragraph{The length of a response.}
The mask could also change the length of a rollout, and the fourth panel of Figure~\otherref{fig:signal}{5} therefore tracks the mean response length of the two runs on \dataset{}. Both of them settle between 600 and 900 tokens once the format is stable, and both of them lengthen again late in training. \methodshort{} gives the longer responses, at 810 tokens against 708 for plain GRPO. The first 4 rollouts of the \methodshort{} run stay at the length of plain GRPO, at 696 tokens, whereas the last 4, which carry the mask in the intervened groups, reach 923 tokens. The extra length therefore comes from the masked rollouts.

\section{Token-level analysis of the mask}
\label{app:drops}

The section above measures the second wave over a whole group, and we now measure it at the level of the token. The top-1 probability exceeds $\tau$ at 2.2\% of the positions the mask examines, as the left panel of Figure~\ref{fig:maskact} shows, and no response reaches the cap on masks of Table~\ref{tab:config}. To see which positions those are, we record every token that the mask forbids together with the token that the sampler draws in its place, over the \methodshort{} run of Table~\otherref{tab:main}{1}. Each of the 8 sampling engines writes the first 20,000 masked positions it produces, which gives 160,000 positions in total, and the sampler resolves every one of them to a replacement. Figure~\ref{fig:dropclouds} shows the two sides of that substitution, and Table~\ref{tab:dropstats} reports where those positions sit and which token the sampler draws in their place.

\begin{figure*}[t]
\centering
\includegraphics[width=\linewidth]{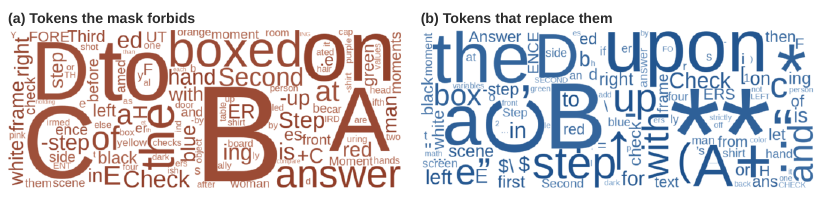}
\caption{The tokens that the mask forbids, and the tokens that the sampler draws in their place, over the masked positions of the \methodshort{} run. Size follows frequency. Both panels fold a token onto its printed form, so a token and its word-boundary variant appear once, and they omit byte pieces that print no full character.}
\label{fig:dropclouds}
\end{figure*}

The two panels of Figure~\ref{fig:dropclouds}, however, do not carry the same kind of word, because the mask forbids mostly the answer letters A to E, \texttt{boxed} and \texttt{answer}, and frequent function words such as \texttt{to} and \texttt{the}. The mask therefore acts most often at the moment the model commits to an answer inside its reasoning span, which is where a near-certain token sits. The tokens \texttt{boxed} and \texttt{answer} appear in this list because the model drafts its answer inside the reasoning span before it closes that span. The mask therefore reaches that draft rather than the answer the reward reads, which lies after the closing think tag and outside the span. The mask never removes the think tags either.

\begin{table*}[t]
\centering
\footnotesize
\begin{tabular}{lll}
\toprule
masked & sampled instead & what the substitution changes \\
\midrule
\multicolumn{3}{l}{\emph{the value that a hop reads}} \\
\texttt{B} & \texttt{C} & the answer the model was about to name \\
\texttt{right} & \texttt{left} & the side of the frame that a spatial hop reads \\
\texttt{bottom} & \texttt{top} & the vertical half that a spatial hop reads \\
\texttt{before} & \texttt{after} & the direction of an order hop \\
\texttt{orange} & \texttt{yellow} & the color that an attribute hop reads \\
\texttt{arms} & \texttt{hands} & the part of a person that a hop refers to \\
\midrule
\multicolumn{3}{l}{\emph{where the rest of the response goes}} \\
\texttt{on} & \texttt{in} & the relation between two objects \\
\texttt{the} & \texttt{left} & a side the model had not yet named \\
\texttt{is} & \texttt{appears} & how firmly the model commits to what it reports \\
\texttt{Second} & \texttt{First} & the step of the chain the model is working on \\
\bottomrule
\end{tabular}

\caption{Notable substitutions of the \methodshort{} run.}
\label{tab:droppairs}
\end{table*}

The substitution also reaches the content of a hop, as Table~\ref{tab:droppairs} shows. The upper block of that table changes the value a hop reads, because the sampler names a different answer letter, one side of the frame for the other, one vertical half for the other, the opposite direction of an order relation, a different color, and a different part of a person. These are the kinds of hop that a \dataset{} question asks about, so the mask reaches the content on which the answer depends and not only the wording that surrounds it. The lower block changes the direction the rest of the response takes, because the sampler alters the relation between two objects, commits to a side that the model had not yet named, weakens the claim the model was about to make, and renames the step of the chain that the model is working on. We therefore treat the second wave as exploration of the answer rather than as noise, and Appendix~\ref{app:signal} supports this at the level of the group, because the mask produces a correct rollout in 0.13 of the all-incorrect groups and gives 0.12 of the groups a solution that the first wave never finds.

\begin{table*}[t]
\centering
\footnotesize
\setlength{\tabcolsep}{5pt}
\begin{tabular}{lc}
\toprule
over the 160,000 masked positions & value \\
\midrule
\multicolumn{2}{l}{\emph{where the mask acts}} \\
median position of a masked token in the response & 242 \\
share within the first 100 tokens of the response & 0.30 \\
share beyond token 400 of the response & 0.37 \\
\midrule
\multicolumn{2}{l}{\emph{how certain the policy is at a masked position}} \\
mean top-1 probability & 0.98 \\
share whose top-1 probability exceeds 0.99 & 0.42 \\
share where the second most likely token holds more than half of the rest & 0.46 \\
\midrule
\multicolumn{2}{l}{\emph{which token the sampler draws instead}} \\
share where the sampler draws the second most likely token & 0.50 \\
share where it draws the second or the third most likely token & 0.65 \\
share where it draws a token outside the 8 that the trace records & 0.18 \\
\bottomrule
\end{tabular}
\caption{The masked positions of the \methodshort{} run, and the tokens the sampler draws in their place.}
\label{tab:dropstats}
\end{table*}

Three rows of Table~\ref{tab:dropstats} matter for the method. First, the mask acts across the whole reasoning span rather than at its opening, because the median masked position sits at token 242 and about one masked position in three sits beyond token 400. Second, the sampler draws the token that the policy itself ranks second at half of the masked positions, and it draws one of the first three candidates at about two thirds of them, so the second wave follows a continuation that the policy already ranks highly rather than an arbitrary token. Third, the second most likely token holds more than half of the probability that the mask leaves behind at 46\% of the positions, hence a masked position is often a choice between the token the policy would commit to and one alternative.

\section{Derivations}
\label{app:deriv}

\paragraph{Entropy at a fixed top-1 probability.}
Let $V$ be the vocabulary size and $\pstar$ the probability of the most likely token. The entropy of the distribution is smallest when one token holds all the remaining mass, whereas it is largest when the mass is spread evenly over the other $V - 1$ tokens, which gives two bounds:
\begin{equation}
\begin{aligned}
H_{\min}(\pstar) ={}& -\pstar \log \pstar - (1 - \pstar)\log(1 - \pstar),\\
H_{\max}(\pstar) ={}& -\pstar \log \pstar \\
&+ (1 - \pstar)\big(\log(V - 1) - \log(1 - \pstar)\big).
\end{aligned}
\label{eq:hbounds}
\end{equation}
With $V = 151{,}936$ the interval is $[0.50, 2.89]$ nats at $\pstar = 0.80$, $[0.33, 1.52]$ at $0.90$, $[0.20, 0.80]$ at $0.95$ and $[0.06, 0.18]$ at $0.99$, so the top-1 probability determines the entropy only within a wide range. In the other direction, a fixed entropy leaves a wide range of top-1 probabilities, because at $H = 0.5$ nats the value of $\pstar$ can lie anywhere in $[0.80, 0.97]$.

\paragraph{Entropy after the mask.}
Because the mask removes the top token and renormalizes the remaining mass, it gives a distribution $q(y) = \pi(y) / (1 - \pstar)$ for $y \ne \ystar$, and the entropy of this distribution has a closed form:
\begin{equation}
H_{\text{post}} = \frac{H + \pstar \log \pstar}{1 - \pstar} + \log(1 - \pstar).
\label{eq:hpost}
\end{equation}
Since a trace stores $\pstar$ to five decimals, we evaluate this form only where $1 - \pstar > 10^{-3}$, below which the denominator loses too much precision.

\paragraph{The ratio at a masked position.}
The mask also changes the importance ratio, because the sampling engine returns the log-probability of each generated token under the distribution it sampled from. At a masked position, that distribution is $\tilde{\pi}_\theta$ of Equation~\otherref{eq:mask}{3}, whose value at the sampled token is $\pi_\theta(o_t \mid \cdot) / (1 - \pstar)$, and the trainer therefore uses this value as the old log-probability. Hence, at the first update, before $\theta$ has moved, the ratio is $\pi_\theta(o_t \mid \cdot) \big/ \big(\pi_\theta(o_t \mid \cdot)/(1 - \pstar)\big) = 1 - \pstar$, which is Equation~\otherref{eq:ratio}{4}. Later updates move the ratio by the same factor as any other token, whereas the policy samples the positions after the mask from $\pi_\theta$ itself, so their ratio is one.

\section{Effect of the second wave on the update}
\label{app:update}

The derivation above concerns one masked position, and we also measure what the second wave does to the update as a whole. Table~\ref{tab:update} compares the two runs of Table~\otherref{tab:main}{1} on four quantities that the trainer logs, over the same window as Table~\ref{tab:signal}, and we again report the mean over the first half of that window and the mean over the second half.

\begin{table}[t]
\centering
\footnotesize
\setlength{\tabcolsep}{3.5pt}
\begin{tabular}{@{}>{\raggedright\arraybackslash}p{3.5cm}cccc@{}}
\toprule
 & \multicolumn{2}{c}{plain GRPO} & \multicolumn{2}{c}{\methodshort{}} \\
\cmidrule(lr){2-3}\cmidrule(lr){4-5}
per training step & early & later & early & later \\
\midrule
share of clipped positions & 0.005 & 0.009 & 0.023 & 0.021 \\
mean magnitude of the advantage & 0.337 & 0.332 & 0.393 & 0.338 \\
gradient norm & 1.23 & 1.30 & 1.26 & 1.16 \\
mean training reward & 0.302 & 0.355 & 0.293 & 0.360 \\
\bottomrule
\end{tabular}
\caption{What the second wave changes in the update, on the two runs that continue on \dataset{}.}
\label{tab:update}
\end{table}

Three of these rows carry the result. First, the clip range catches more positions under \methodshort{} than under plain GRPO, at 0.023 and 0.021 against 0.005 and 0.009, and we attribute this to the forced tokens of the second wave, because a forced token starts at a ratio of $1 - \pstar_t \leq 0.05$, so the clip gives it no gradient whenever its advantage is negative (Appendix~\ref{app:deriv}). However, that share stays near two positions in a hundred, so the clip absorbs the perturbation rather than removing the rollout that carries it. Second, the magnitude of the advantage is larger under \methodshort{} early in training, and we attribute this to the restored groups, because a group whose rewards vary gives a non-zero advantage to every one of its rollouts. Third, the two runs reach the same training reward, at 0.360 against 0.355 in the second half of the window, so \methodshort{} obtains the extra groups of Table~\ref{tab:signal} at the same training reward as plain GRPO. In addition, the gradient norm stays within 0.14 of plain GRPO over the whole window, hence the second wave does not change the scale of the update.

\section{Computational cost and response length}
\label{app:cost}

We measure how much time the second wave costs, and how the mask changes the length of a response, on the two runs of Table~\otherref{tab:main}{1} that continue on \dataset{}, namely plain GRPO and \methodshort{}, which we train on the same 4 nodes.

\begin{table}[t]
\centering
\footnotesize
\setlength{\tabcolsep}{4pt}
\begin{tabular}{@{}lrrr@{}}
\toprule
per training step & plain GRPO & \methodshort{} & difference \\
\midrule
step time (s) & 2,275 & 2,285 & $+0.4\%$ \\
wall-clock between steps (s) & 2,435 & 2,459 & $+1.0\%$ \\
generation (s) & 1,097 & 1,084 & $-1.2\%$ \\
actor update (s) & 1,167 & 1,190 & $+2.0\%$ \\
advantage and check (s) & 0.15 & 0.58 & $+0.4$ s \\
tokens per second & 281 & 285 & $+1.4\%$ \\
actor model utilization & 0.137 & 0.139 & --- \\
trainer idle ratio & 0.48 & 0.47 & --- \\
\bottomrule
\end{tabular}
\caption{Training cost of the two runs that continue on \dataset{}, averaged over the run, on 4 nodes of 8 H100 GPUs. The wall-clock row is the mean gap between the timestamps of consecutive steps.}
\label{tab:cost}
\end{table}

\paragraph{Training time.}
Table~\ref{tab:cost} reports the timing that the trainer logs, averaged over the run. Every timing row of that table, except the advantage and check step, differs by at most 2.0\% between the two runs, and the largest of these differences is the actor update, which the \methodshort{} run makes slower because this run produces slightly longer sequences. The advantage and check step itself grows from 0.15 to 0.58 seconds, which is negligible compared with a step of more than 2,000 seconds. In addition, token throughput differs by 1.4\%, model utilization is 0.137 against 0.139, and the trainer waits for rollouts during about 48\% of its time in both runs, so neither run is limited by the extra computation that \methodshort{} adds. This result follows from the design of \methodshort{}, because we keep the compute budget fixed at 8 rollouts per question and we apply the mask inside the sampler as a logits processor, so the second wave adds no forward passes beyond the ones that plain GRPO already spends. Although the second wave waits for the rewards of the first wave, the asynchronous rollouter samples many other questions during that wait, so we measure no cost from this wait in the step time.

\paragraph{Response length.}
The \methodshort{} run gives longer responses on average, at 810 tokens against 708 for plain GRPO, and Appendix~\ref{app:signal} shows that the extra length comes from the rollouts that carry the mask. In addition, fewer than 0.4\% of the responses of either run reach the 16,384-token limit, so truncation costs neither run a measurable amount of reward.

\fi

\end{document}